\documentclass[sigconf]{acmart}

\AtBeginDocument{%
  \providecommand\BibTeX{{%
    \normalfont B\kern-0.5em{\scshape i\kern-0.25em b}\kern-0.8em\TeX}}}

\copyrightyear{2020}
\acmYear{2020}
\setcopyright{acmlicensed}
\acmConference[MM '20]{Proceedings of the 28th ACM International Conference on Multimedia}{October 12--16, 2020}{Seattle, WA, USA}
\acmBooktitle{Proceedings of the 28th ACM International Conference on Multimedia (MM '20), October 12--16, 2020, Seattle, WA, USA}
\acmPrice{15.00}
\acmDOI{10.1145/3394171.3413689}
\acmISBN{978-1-4503-7988-5/20/10}

\usepackage{epstopdf}
\usepackage{multirow}
\usepackage{flushend}

\clearpage
\begin{document}
\fancyhead{}
\title{Hierarchical Bi-Directional Feature Perception Network for Person Re-Identification}


\author{Zhipu Liu}

\affiliation{
  \institution{School of Microelectronics and Communication Engineering, Chongqing University}
 \streetaddress{Shazheng street No.174, Shapingba District}
  \city{Chongqing}
  \state{China}
  \postcode{400044}
}
\email{zpliu@cqu.edu.cn}

\author{Lei Zhang}
\authornote{Corresponding author}

\affiliation{
  \institution{School of Microelectronics and Communication Engineering, Chongqing University}
 \streetaddress{Shazheng street No.174, Shapingba District}
  \city{Chongqing}
  \state{China}
  \postcode{400044}
}
\email{leizhang@cqu.edu.cn}

\author{Yang Yang}

\affiliation{
  \institution{School of Computer Science and Engineering, University of Electronic Science and Technology of China}
 \streetaddress{2006 Xiyuan Ave, West Hi-Tech Zone}
  \city{Chengdu}
  \state{China}
  \postcode{611731}
}
\email{dlyyang@gmail.com}




\begin{abstract}
Previous Person Re-Identification (Re-ID) models aim to focus on the most discriminative region of an image, while its performance may be compromised when that region is missing caused by camera viewpoint changes or occlusion. To solve this issue, we propose a novel model named Hierarchical Bi-directional Feature Perception Network (HBFP-Net) to correlate multi-level information and reinforce each other. First, the correlation maps of cross-level feature-pairs are modeled via low-rank bilinear pooling. Then, based on the correlation maps, Bi-directional Feature Perception (BFP) module is employed to enrich the attention regions of high-level feature, and to learn abstract and specific information in low-level feature. And then, we propose a novel end-to-end hierarchical network which integrates multi-level augmented features and inputs the augmented low- and middle-level features to following layers to retrain a new powerful network. What's more, we propose a novel trainable generalized pooling, which can dynamically select any value of all locations in feature maps to be activated. Extensive experiments implemented on the mainstream evaluation datasets including Market-1501, CUHK03 and DukeMTMC-ReID show that our method outperforms the recent SOTA Re-ID models.

\end{abstract}





\maketitle

\begin{figure}[t]
\begin{center}
\includegraphics[width=1.0\linewidth]{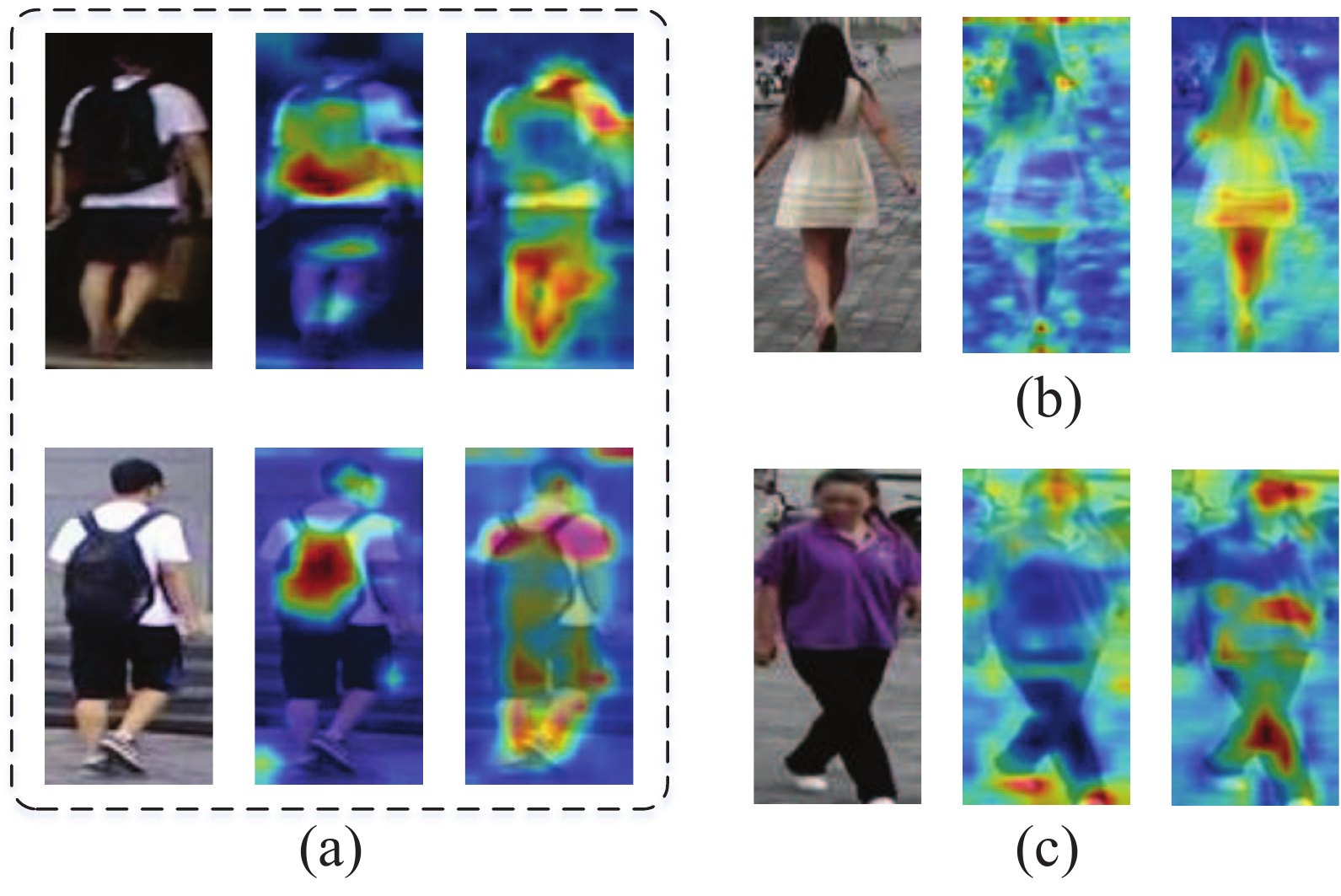}
\end{center}
   \caption{The visualization \cite{Zhou_2016_CVPR} of high- and low-level feature maps. In each example, from left to right are the original image, the visualization of original features and augmented features. (a) Images from two different identities and their visualization of high-level feature maps. (b)-(c) Two visual examples of original and augmented low-level feature maps.   }
\label{fig1}
\end{figure}
\section{INTRODUCTION}
Person re-identification (Re-ID) aims to retrieve a query person from all the gallery images captured by different cameras without view overlap. With the explosion of Convolutional Neural Networks(CNN), lots of deep learning based methods \cite{Sun_2018_ECCV, Si_2018_CVPR, Zhous_2019_ICCV, DBLP:journals/corr/HermansBL17, Xu_2018_CVPR, 8607050, wang2020self, yang2018person} have achieved significant breakthrough in person Re-ID task. However, due to various challenges such as misaligned, occlusion, background clutter and pose changes, this task remains an unsolved problem.

Due to the weak diversity of person Re-ID training datasets, the Re-ID models tend to focus only on the most discriminative regions of an image, while some local details and non-salient parts can be easily ignored during feature learning procedure, which may result in low-generalization of the learned features \cite{Wang_2018_acm, Hao_2018_CVPR, Dai_2019_ICCV}. For example, the both images in Fig.~\ref{fig1} (a) carry the same style of backpack, which attract the most attention shown in the second column, while local details like shoes and trousers are ignored. The person Re-ID task suffers from many challenges such as view changes or occlusion, which may cause the attentive discriminative regions to disappear in the desired image. For example, the backpack appears in the back view in Fig.~\ref{fig1} (a), while it is missing when view changes to front. In this situation, non-salient parts and local details, such as shoes and trousers, become key factors to identify the desired image. What's more, when two pedestrian images of different identities contain same salient region, such as backpack in Fig.~\ref{fig1} (a), non-salient parts and local details can be essential information to discriminate difference between them. Therefore, the Re-ID model should focus on the overall feature patterns to learn more reliable features and make the model more robust to partial occlusion or camera viewpoint changes.

To solve this problem, part-based methods \cite{Sun_2018_ECCV, Wang_2018_acm, Yang_2018_AAAI, Zheng_2019_CVPR} were proposed, which split the feature maps into several uniform strips to force each partition stripe to meet an individual ID-prediction loss for enriching salient regions. Dropout-based method is another effective way to enlarge attention parts by erasing or dropping some regions, which can force the network to focus on the remaining regions. However, the performance of part-based approaches heavily rely on the employed partition mechanism, which can be comprised when misalignment or occlusion arise. Dropout-based methods \cite{Dai_2019_ICCV, DBLP:journals/corr/abs-1708-04896, Xia_2019_ICCV} in some way relieved this dilemma, but pre-defined erased regions are hard to confirm and vary in different datasets, which may not generalize well.

Many previous works \cite{Wang_2018_CVPR, Chang_2018_CVPR} directly employ loss function on low-level features to solve the problem that the supervision signal from high level loss is indirect and weak for low-level layers, the information of low-level features, however, is dispersive, which is unable to apply distinctive identity representation of a pedestrian, so lower layers are very easily misguided if low-level features are not augmented to obtain abstract information. For example, two visual examples of original low-level feature maps shown in the second column in Fig.~\ref{fig1} (b) and (c), it can be observed that the low-level feature maps without augmentation can't capture discriminative semantic regions and the high response areas scatter to full image. Considering that the features from shallow layers are prone to capture dispersive local details and those from deep layers tend to focus on the discriminative semantic regions, MDA \cite{Liu_2017_ICCV} uses attention masks to correlate and augment each other, the masks, however, learned from lower layers are usually in poor quality \cite{Zhou_2019_ICCV}. Previous works neglect that the features extracted from different layers are mutually correlated and can reinforce each other.

Motivated by these observation, in this paper, we propose a novel Hierarchical Bi-directional Feature Perception Network (HBFP-Net), which exploits the relationship of multi-level feature maps via low-rank bilinear pooling and then Bi-directional Feature Perception(BFP) module is employed to correlate multi-level information. Through BFP, the deep layer features, augmented by shallow layer features, can enlarge the attention to non-salient and local regions, and in the same way, the lower features, augmented by higher features, can concentrate attention on discriminative semantic regions and learn more abstract and specific information that in turn benefit to the network. As illustrated in Fig.~\ref{fig1}, the third column images of (a), (b) and (c) represent the augmented higher and lower level features. It can be observed that the augmented high-level features enlarge the attention to whole body and the information of augmented low-level features can be concentrated on the main parts of body. For further enriching the representation capability of final features, the reinforced features are input to the following layers to retrain a more powerful network. Note that all of these manipulations are end-to-end process.

Max and average pooling are widely used in person Re-ID due to its super performance in capturing the most discriminative and global information, some sub-important information, however, is excluded from max pooling and suppressed in average pooling. Since the learned features augmented by BFP module enlarge the attention regions to whole body, non-salient parts and local details also attract much attention in learned feature maps, so how to preserve these gradual cues during pooling precess is very important for the final descriptor of a pedestrian image. To solve this problem, we propose a novel generalized pooling to capture not only the most discriminative parts but also some detailed gradual cues, which can dynamically select any value of all locations in feature maps to be activated rather than the largest response value or global information including background.

\begin{figure*}
\includegraphics[width=1.0\linewidth]{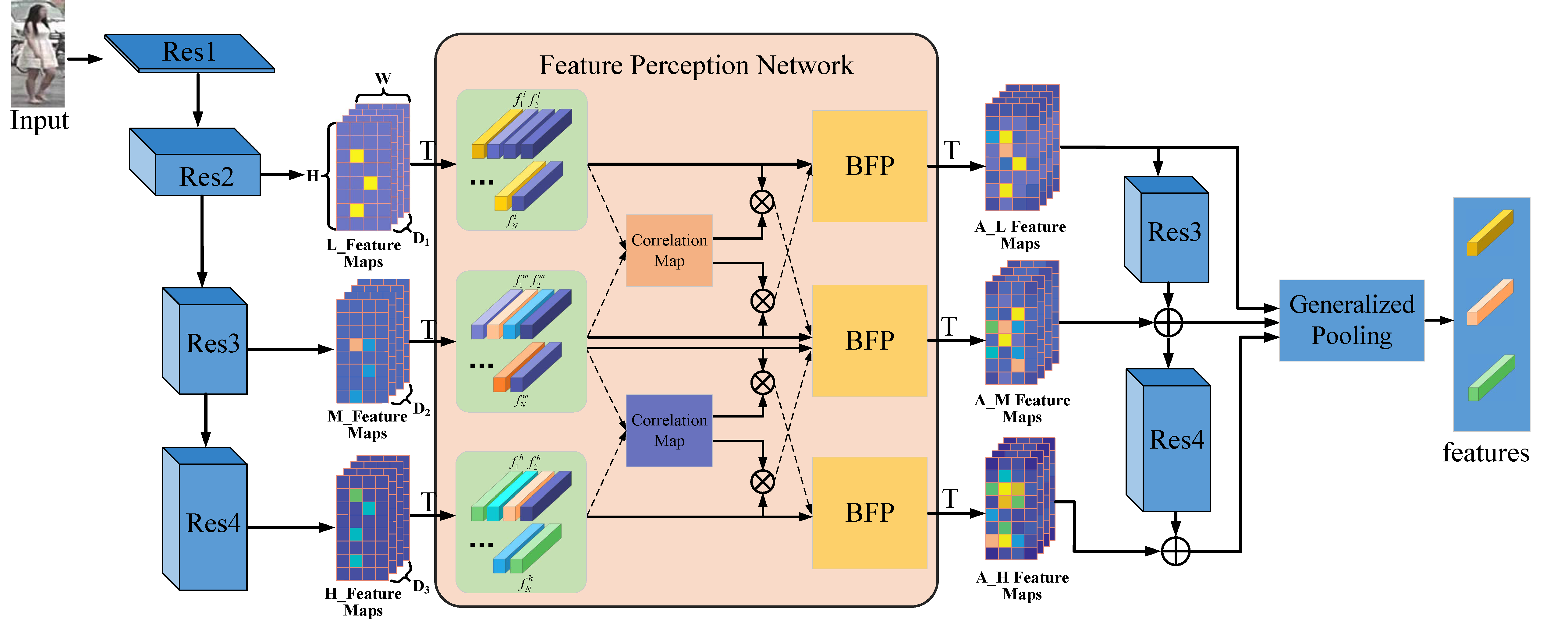}
   \caption{Structure of HBFP-Net. The input is the feature maps extracted from the res\_conv4\_1, res\_conv4 and res\_conv5 of ResNet-50 as low-, middle- and high-level features. `T' and $ \otimes $ represent transpose operation and matrix multiplication. During training, triplet loss and Cross-Entropy loss are employed to each learned feature. During testing, these three features are concatenated as the final descriptor of a pedestrian image.}
\label{fig2}
\end{figure*}

\section{RELATED WORKS}
With the development of deep learning methods, great progress have been achieved in person Re-ID task. Among all these methods, part and attention based methods \cite{Sun_2018_ECCV, Yang_2018_AAAI, Zheng_2019_CVPR, Si_2018_CVPR, Li_2018_CVPR, Suh_2018_ECCV} are two main ways to learn discriminative features and improve the retrieval performance. Beside these two methods, fusion of multi-level information \cite{Liu_2017_ICCV, Wang_2018_CVPR, Chang_2018_CVPR} and dropout based methods \cite{DBLP:journals/corr/abs-1708-04896, Dai_2019_ICCV, Xia_2019_ICCV} also achieved superior performance.

To capture the most salient parts, attention mechanism is widely used to CNN and achieved promising performance. \cite{Zhao_2017_ICCV} proposed a part-aligned network, which can learn aligned discriminative parts automatically in an unsupervised manner. In \cite{Chen_2019_ICCV}, both channel and spatial attention were used and a regularization was introduced to learn both salient and non-salient parts. \cite{Chens_2019_ICCV} proposed a high-order attention module to produce more powerful attention masks. The attention mechanism often enforce the features to capture the most discriminative parts, while some non-salient regions may be ignored \cite{Chen_2019_ICCV}. To relieve this dilemma, many part and dropout based methods were proposed. The part based approaches force the network to learn local discriminative regions from each partition strip and dropout based methods focus on the remaining regions after random dropping a square area.

\cite{Sun_2018_ECCV} proposed a typical part method and a strong baseline, which splits the last feature maps into several uniform strips to capture local details. Based on this work, lots of carefully designed part based models were proposed. In \cite{Zheng_2019_CVPR}, pyramid based network was proposed to capture coarse-to-fine information through various partition scales. Another more intuitive approach is dropout based methods, which randomly drop a square area in feature maps or input images to force the network to focus on the remaining regions \cite{DBLP:journals/corr/abs-1708-04896, Dai_2019_ICCV, Xia_2019_ICCV}. \cite{Dai_2019_ICCV} proposed a Batch DropBlock (BDB) network, which randomly drops the same region of feature maps in a batch to force the attentive feature to focus on the remaining regions. Based on this work, \cite{Xia_2019_ICCV} proposed a second-order non-local attention method to learn non-local attention masks through encoding location-to-location feature level relationships.

To capture various semantic information, \cite{Wang_2018_CVPR} trains multi-level semantic features individually and uses weighted sum to get final feature for retrieval. \cite{Chang_2018_CVPR} proposed a Multi-Level Factorisation Network, which contains a set of factor modules and a factor selection module, and the factor selection module can dynamically select any factor module to be activated. \cite{Liu_2017_ICCV} proposed a HydraPlus-Net (HP-net), which learns attention maps of multiple intermediate layers and multi-directionally feed these masks to different feature layers to enrich the final feature representation. \cite{Zhou_2019_ICCV} proposed a consistent attention regularizer to keep the learned attention masks similar from low-level, mid-level and high-level feature maps. These methods use attention masks to correlate multi-level information, the quality of masks deduced from low-level features, however, may be poor for that the information in low layer features is too dispersed and fail to concentrate attention on discriminative regions. Previous works neglect that various semantic features are mutually correlated and can reinforce each other. So we exploit the relationship of multi-level features by low-rank bilinear pooling \cite{DBLP:journals/corr/KimOKHZ16} and reinforce each other through bi-directional feature perception module to enlarge the attention to non-salient and local regions of high-level features and learn more abstract and specific information in low-level features that in turn benefits the deep network.

\section{Proposed Method}
In this section, we will present the structure of HBFP-Net shown in Fig.~\ref{fig2}. And details of the BFP and proposed generalize pooling are described in section 3.2 and 3.3.

\subsection{Network Architecture Overview}
The HBFP-Net can take any deep network, such as Google-Net \cite{Szegedy_2015_CVPR}, Densenet \cite{DBLP:journals/corr/IandolaMKGDK14} and ResNet \cite{He_2016_CVPR}, as backbone. Our paper choose the ResNet-50 as backbone due to its powerful feature representation ability for person Re-ID \cite{Chens_2019_ICCV, Zhengs_2019_CVPR, Wang_2018_acm, Luo_2019_Strong_TMM, DBLP:journals/corr/abs-1711-08184}. To make the backbone more suitable for person Re-ID task, the last spatial down-sampling operation between res\_conv4 and res\_conv5 is removed to get larger size of feature maps \cite{Sun_2018_ECCV, Zhang_2019_CVPR, Chen_2019_ICCV, Dai_2019_ICCV}.

ResNet-50 contains four blocks and each block comprises of multiple convolutional layers. The feature maps extracted from the second, third and fourth block are used as low-, middle- and high-level semantic features, which are expressed as L\_Feature Maps, M\_Feature Maps and H\_Feature Maps. The structure of HBFP-Net is shown in Fig.~\ref{fig2}. The three feature maps are 3-dimensional tensors with the size of $D \times H \times W$, which represent the number of channel, height and weight, respectively. Then the feature maps are reshaped to 2-dimensional tensors of size $D \times N$, where $N = H \times W$, which can be viewed as N local features and each local feature is a $D$-dimension vector. Based on each local feature-pair of different level feature maps, the correlation maps ${C_{(L,M)}}$ and ${C_{(M,H)}}$ are modeled via low-rank bilinear pooling, which represent the relationship of L\_Features Maps \& M\_Features Maps and M\_Features Maps \& H\_Features Maps, respectively. And then multi-level feature maps correlate each other through information exchange. For example, the matrix multiplication of L\_Feature Maps and ${C_{(L,M)}}$ will map the information of low-level feature to middle level, which can be expressed as $B(M,L \otimes {C_{(L,M)}})$, where $L$ and $M$ represent low- and middle-level feature maps. So the process of bi-directional feature perception of multi-level features can be expressed as $B(L,M \otimes {C_{(L,M)}})$, $B(M,L \otimes {C_{(L,M)}})$, $B(M,H \otimes {C_{(M,H)}})$ and $B(H,M \otimes {C_{(M,H)}})$, where the $B()$ represents BFP module. Note that the middle-level feature maps augmented by low- and high-level feature maps will be summed as the augmented feature. The details of correlation map and BFP module will be described in next section.

The output of BFP module, followed by transpose operation, are three augmented feature maps, represented as A\_L Feature Maps, A\_M Feature Maps and A\_H Feature Maps. We argue that the augmented features contain multi-level fusion information, which can offer more powerful representation. So we input A\_L Feature Maps and A\_M Feature Maps to following blocks to integrate multi-level information in network. Specifically, the A\_L Feature Maps are fed into block3 and the sum of its output and A\_M Feature Maps is input to block4 to learn an integration feature maps of I\_Feature Maps. To integrate multi-level information, we sum the I\_Feature Maps and A\_H Feature Maps as the fusion feature. In this way, three augmented feature maps with integrated information are learned, each of which contains not only multi-level semantic information but also global and local details. To fully exploit the rich information, these three features are followed by generalized pooling, which can not only obtain the most discriminative and global information, but also capture the detailed gradual cues between them.

\subsection{Feature Perception Network}
Let X $ \in {R^{{D_1} \times {H_1} \times {W_1}}}$, Y $ \in {R^{{D_2} \times {H_2} \times {W_2}}}$ and Z $ \in {R^{{D_3} \times {H_3} \times {W_3}}}$ represent the feature maps extracted from the low-, middle- and high-level layers. Note that to reduce the computation complexity, the low-level feature maps here are extracted from res\_conv4\_1 and since the last down-sampling operation between res\_conv4 and res\_conv5 is removed, so the X, Y and Z have same spatial size ($H = {H_1} = {H_2} = {H_3}$, $W = {W_1} = {W_2} = {W_3}$). Essentially, the extracted feature maps can be divided into $H \times W$ local regions, and each local region is a 2-D array of D-dimensional(${D_1}$, ${D_2}$ or ${D_3}$) local feature. Therefore, it can be viewed as $H \times W$ local features expressed as $X = \left\{ {{x_i}\left| {i = 1,2, \cdot  \cdot  \cdot ,N} \right.} \right\}$, $Y = \left\{ {{y_i}\left| {i = 1,2, \cdot  \cdot  \cdot ,N} \right.} \right\}$ and $Z = \left\{ {{z_i}\left| {i = 1,2, \cdot  \cdot  \cdot ,N} \right.} \right\}$, where $N = H \times W$. For clarity, we use X and Y to describe the detailed process of BFP.

Before bi-directional feature perception manipulation, we first re-weight each location of spatial regions via simple self-awareness operation. For feature maps $X$ and $Y$, a $1 \times 1$ convolutional layer is utilized to learn 2-dimensional masks ${M_x}$ and ${M_y}$ and then applied to the origin feature maps:
\begin{equation}
\label{E1}
{\begin{array}{l}
X' = {M_x} \odot X\\
Y' = {M_y} \odot Y
\end{array}}
\end{equation}
where $\odot$ represents element-wise product.
\begin{figure}[t]
\begin{center}
\includegraphics[width=1.0\linewidth]{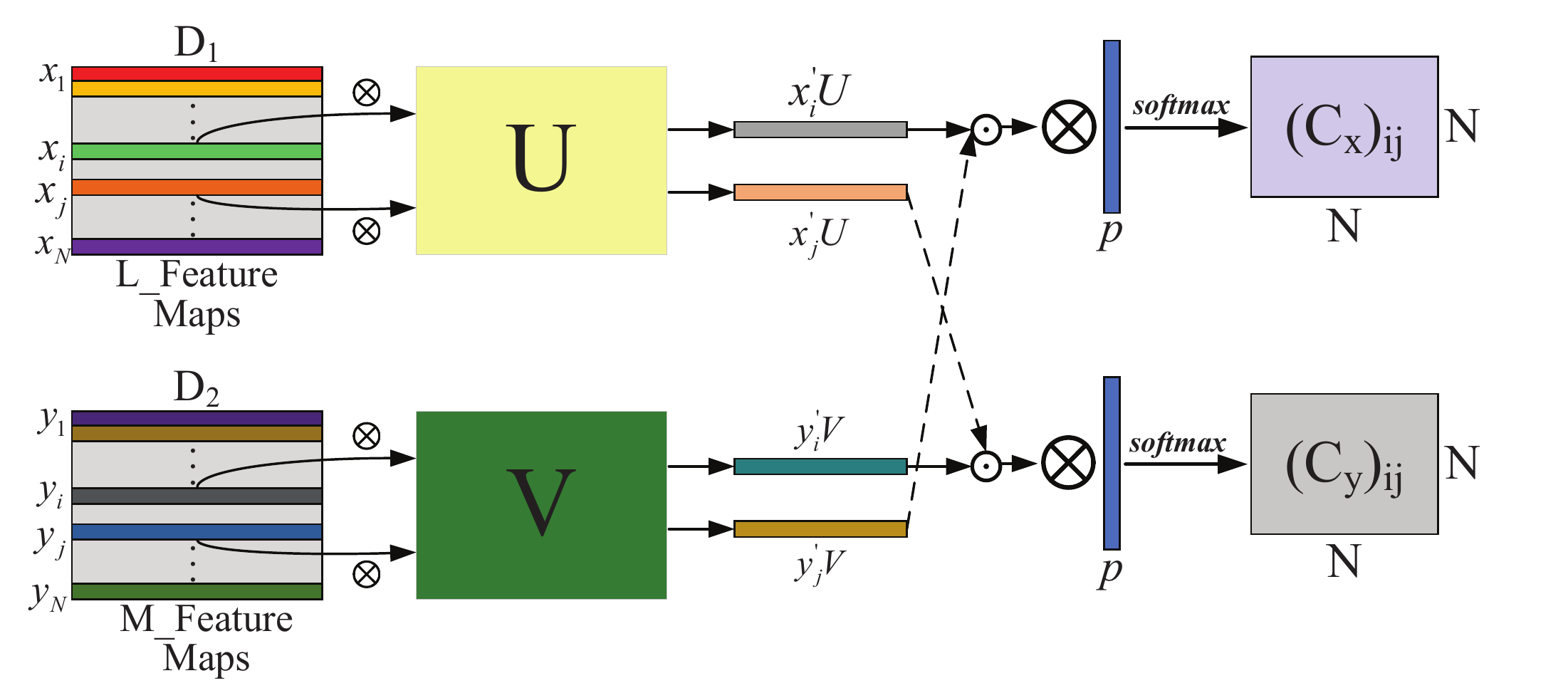}
\end{center}
   \caption{Process of learning correlation map of low- and middle-level feature maps. Softmax function is employed to each column of correlation map ${C_x}$ and ${C_y}$, and ${C_y}$ is the transpose of ${C_x}$.}
\label{fig3}
\end{figure}

\textbf{Correlation maps}. Correlation maps aim to represent the relationship between different level feature maps, which provide the representation of correlation distribution by considering each local feature-pair of feature maps, e.g. ${C_{i,j}}$ represents the relationship between the $i$-th and $j$-th local feature of $X'$ and $Y'$. In this way, the correlation maps of different level features can be confirmed by considering all local feature-pairs.

The detailed process of correlation maps learning is shown in Fig.~\ref{fig3}. We first rearrange the local features of feature maps $X'$ and $Y'$ into a matrix form by stacking each local representation ${x'_i}$ and ${y'_j}$ in row direction expressed as $X' = [x_1', \cdot  \cdot  \cdot ,x_i', \cdot  \cdot  \cdot ,x_N'] \in {R^{N \times {D_1}}}$ and $Y' = [y_1', \cdot  \cdot  \cdot ,y_j', \cdot  \cdot  \cdot ,y_N'] \in {R^{N \times {D_2}}}$, where $N = H \times W$. And the relationship of local feature $x_i'$ and $y_j'$ can be expressed as:
\begin{equation}
\label{E2}
{{C_{i,j}} = {p^T}(\sigma ({x_i'}{U}) \odot \sigma ({y_j'}{V}))}
\end{equation}
where $ \odot $ and $\sigma $ represents Hadamard product (element-wise multiplication) and ReLU \cite{Vinod_2010_ICML} non-linear activation function, respectively. $U$ and $V$ are two projection matrix, which aim to map the feature maps $X'$ and $Y'$ to pooling space, where $U \in {R^{{D_1} \times L}}$, $V \in {R^{{D_2} \times L}}$ and $L$ is the projection dimension. $p$ is the linear vector in low-rank bilinear pooling, where $p \in {R^L}$. To obtain correlation map $C$, the above operations can be rewritten as a matrix form:
\begin{equation}
\label{E3}
{C_x = (({\rm{I}} \cdot {p^T}) \odot \sigma (X'U)) \cdot \sigma ({V^T}{(Y')^T}))}
\end{equation}
where $I \in {R^N}$ and $C \in {R^{N \times N}}$. $\odot$ and $\cdot$ represent element-wise multiplication and matrix multiplication. $C_x$ represents the correlation distribution of $X'$ when map the information of $X'$ to $Y'$. In the same way, $C_y$ represents the correlation distribution of $Y'$ and $C_y$ is the transpose of $C_x$. Then $softmax$ function is applied to each columns of $C_x$ and $C_y$. Both $C_x$ and $C_y$ are expressed as ${C_{(L,M)}}$ in section 3.1.

\begin{figure}[t]
\begin{center}
\includegraphics[width=1.0\linewidth]{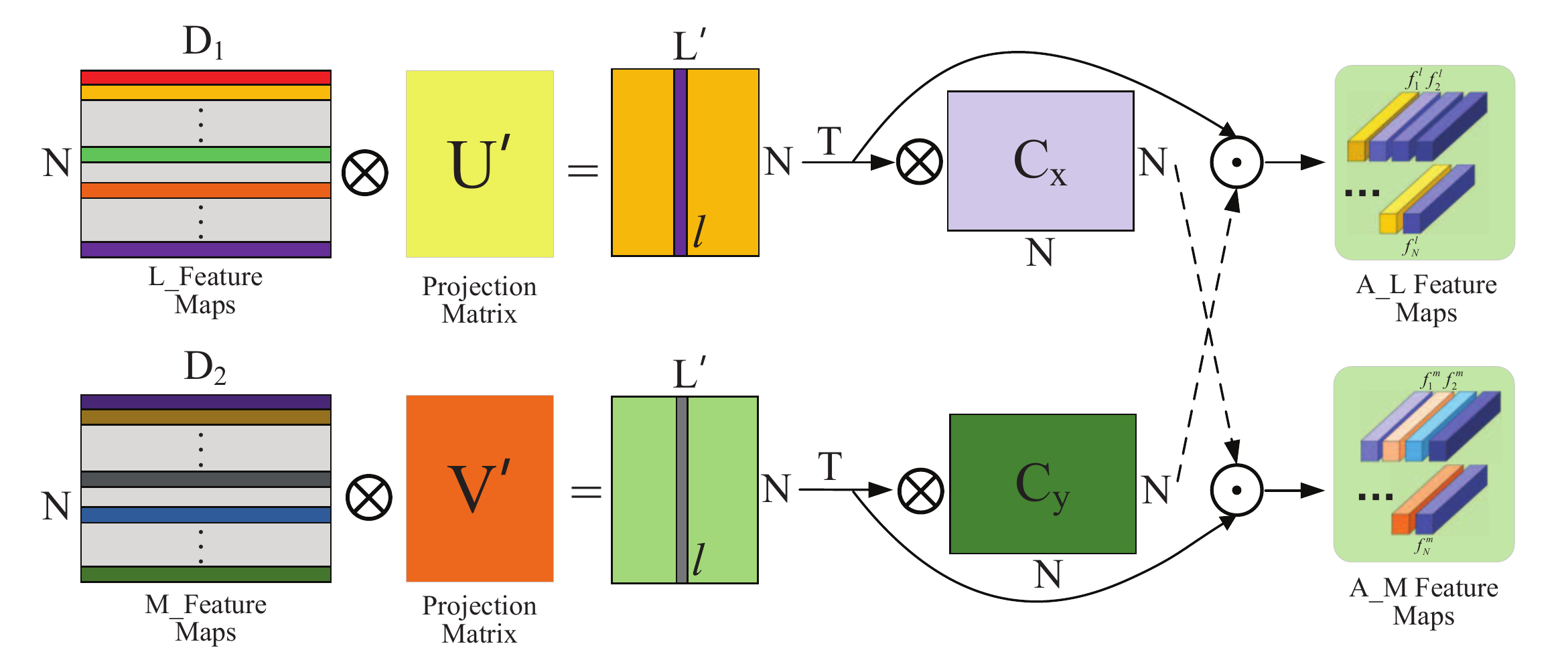}
\end{center}
   \caption{Structure of BFP module. The input is low- and middle-level feature maps. `T' represents transpose operation. $ \otimes $ and $ \odot $ represent matrix multiplication and Hadamard product.}
\label{fig4}
\end{figure}

\textbf{Bi-directional Feature Perception}. Based on the correlation maps ${C_x}$ and ${C_y}$, we correlate multi-level information via bi-directional feature perception manipulation, and the details are shown in Fig.~\ref{fig4}. The low- and high-level feature maps $X'$ and $Y'$ are first mapped to pooling space by projection matrix $U'$ and $V'$. And then the two feature maps on pooling space correlate and reinforce each other through corresponding correlation maps. The above operations can be written as:
\begin{equation}
\label{E4}
{\begin{array}{l}
{({X_y})_l} = {(\sigma {(X'U')_l})^T} \odot ({(\sigma {(Y'V')_l})^T} \cdot {C_y})\\
{({Y_x})_l} = ({(\sigma {(X'U')_l})^T} \cdot {C_x}) \odot {(\sigma {(Y'V')_l})^T}
\end{array}}
\end{equation}
where $U' \in {R^{{D_1} \times L'}}$ and $V' \in {R^{{D_2} \times L'}}$ are linear mappings. ${(X'U')_l} \in {R^N}$ and ${(Y'V')_l} \in {R^N}$. ${C_x}$ and ${C_y}$ are corresponding correlation maps learned from Eq.~(\ref{E3}). ${{({X_y})_l} \in {R^N}}$ and ${{({Y_x})_l} \in {R^N}}$, which denote the $l$-th group elements of the augmented low- and middle-level feature maps $X_y$ and $Y_x$. The subscript $l$ is the index of column. $X_y \in {R^{L' \times N}}$ represents the information of the middle-level feature maps $Y'$ mapped to low-level feature maps $X'$ via the corresponding correlation maps ${C_y}$ and the augmented middle-level feature maps $Y_x$ has the same way.
Finally, the bi-directional perception features ${{\tilde X}_y}$ and ${{\tilde Y}_x}$ can be obtained:
\begin{equation}
\label{E5}
{\begin{array}{*{20}{l}}
{{{\tilde X}_y} = {{({X_y})}^T}{P_x}}\\
{{{\tilde Y}_x} = {{({Y_x})}^T}{P_y}}
\end{array}}
\end{equation}
where ${{P_x} \in {R^{L' \times D}}}$ and ${{P_y} \in {R^{L' \times D}}}$, which are learnable projection matrix and $D$ is the dimension of the final bi-directional perception features.

\subsection{Generalized Pooling}
Average and max pooling are widely used in person Re-Id for its strong ability to perceive the global spatial information and most discriminative regions, respectively. However, the global information perceived by the average pooling contains not only the whole-body appearance but also the background context. And the max pooling only preserves the largest response values of all spatial locations, while some weak but useful information is ignored. Thus, we use generalized pooling to capture the detailed gradual cues between them.

Specifically, the bi-directional perception feature maps ${{\tilde X}_y}$ and ${{\tilde Y}_x}$ are first normalized to between zero and one, expressed as ${N_x}$ and ${N_y}$. For clarity, we just use ${{\tilde X}_y}$ to describe the detailed process. A pre-set threshold $\lambda $ is used to select those that are bigger than it, and the remaining values are set to zero, which can be written as:
\begin{equation}
\label{E6}
{{({X_y})_G} = \left\{ {\begin{array}{*{20}{c}}
{{{({{\tilde X}_y})}_{i,j}},}\\
{0,}
\end{array}\begin{array}{*{20}{c}}
{}\\
{}
\end{array}} \right.\begin{array}{*{20}{c}}
{{{({N_x})}_{i,j}} \ge \lambda }\\
{{{({N_x})}_{i,j}} < \lambda }
\end{array}}
\end{equation}
where ${{{({{\tilde X}_y})}_{i,j}}}$ and ${{{({N_x})}_{i,j}}}$ denote the element in the $i$-th row and $j$-column of ${{{\tilde X}_y}}$ and ${{N_x}}$. Then just average those preserved values, which can not only excludes background noise but also learn more important information. Generalized pooling is crucial for our learned bi-directional perception feature maps for the reason that the augmented feature maps enlarge attention area to whole body rather than the most discriminative regions, so exploitation of detailed gradual cues can be effective to offer rich information. It can be observed that the average and max pooling are special cases of generalized pooling ($\lambda  = 0$ and $\lambda  = 1$). Thus, we can select any gradual cue by different setting of $\lambda $.

\subsection{Loss Function}
In our experiment, we use widely-used batch-hard triplet loss \cite{DBLP:journals/corr/HermansBL17} and cross entropy loss \cite{Szegedy_2016_CVPR} to train each learned augmented feature. Given a batch of images $X$, consisting of $P$ individuals and $K$ images of each identity, and triplet loss is computed as:
\begin{equation}
\label{E7}
\begin{aligned}
{\mathcal{L}_{triplet}} &=  {\sum\limits_{p = 1}^P\sum\limits_{k = 1}^K {{{\left[ {{d_{pos}^{p,k}}  - {{d_{neg}^{p,k}} }  + m} \right]}_ + }} }\\
\end{aligned}
\end{equation}
where $d_{pos}^{p,k}$ and $d_{neg}^{p,k}$ represent the Euclidean distance of hard positives and negative, respectively. ${\left[  \cdot  \right]_ + } = \max (0, \cdot )$ and $m$ is a margin that controls the distance between positives and negatives.

\begin{table*}
\caption{Comparison (\%) of our method with state-of-the arts on Market-1501, DukeMTMC-ReID and CUHK03-NP (labeled and detected).}
\setlength{\abovecaptionskip}{0.cm}
\setlength{\belowcaptionskip}{-0.cm}
\begin{center}
\begin{tabular}{c|c|c|c|c|c|c|c|c|c}
\hline
\multirow{3}{*}{Methods} & \multirow{3}{*}{Model} & \multicolumn{2}{|c|}{\multirow{2}{*}{Market-1501}} & \multicolumn{2}{|c|}{\multirow{2}{*}{DukeMTMC-ReID}} & \multicolumn{4}{|c}{CUHK03-NP}\\
\cline{7-10}
& \multicolumn{1}{|c|}{} & \multicolumn{1}{|c}{} & \multicolumn{1}{c|}{} & \multicolumn{1}{|c}{} & \multicolumn{1}{c|}{} & \multicolumn{2}{|c|}{Labeled} & \multicolumn{2}{|c}{Detected}\\
\cline{3-10}
&\multirow{3}{*}{} & Rank-1 & mAP & Rank-1 & mAP & Rank-1 & mAP & Rank-1 & mAP\\
\hline
HA-CNN \cite{Li_2018_CVPR} (CVPR2018) & HA-CNN & 91.2 & 75.7 & 80.5 & 63.8 & 44.4 & 41.0 & 41.7 & 38.6\\
MLFN \cite{Chang_2018_CVPR} (CVPR2018) & MLFN & 90.0 & 74.3 & 81.0 & 62.8 & 54.7 & 49.2 & 52.8 & 47.8\\
Mancs \cite{Wang_2018_ECCV} (ECCV2018) & ResNet-50 & 93.1 & 82.3 & 84.9 & 71.8 & 69.0 & 63.9 & 65.5 &  60.5\\
AANet-152 \cite{Tay_2019_CVPR} (CVPR2019) & ResNet-152 & 93.9 & 83.4 & 87.7 & 74.3 & - & - & - & -\\
MGCAM \cite{Song_2018_CVPR} (CVPR2018) & MSCAN \cite{Li_2017_CVPR} & 83.8 & 74.3 & - & - & 50.1 & 50.2 & 46.7 & 46.9\\
SCAL \cite{Chenss_2019_ICCV} (ICCV2019) & ResNet-50 & 95.8 & 89.3 & 89.0 & 79.6 & 74.8 & 72.3 & 71.1 & 68.6\\
\hline
PCB+RPP \cite{Sun_2018_ECCV} (ECCV2018) & ResNet-50 & 93.8 & 81.6 & 83.3 & 69.2 & - & - & 63.7 & 57.5\\
HPM \cite{Yang_2018_AAAI} (AAAI2018) & ResNet-50 & 94.2 & 82.7 & 86.6 & 74.3 & - & - & 63.9 & 57.5\\
MGN \cite{Wang_2018_acm} (ACM MM2018) & ResNet-50 & 95.7 & 86.9 & 88.7 & 78.4 & 68.0 & 67.4 & 66.8 & 66.0\\
Pyramid-Net \cite{Zheng_2019_CVPR} (CVPR2019) & ResNet-50 & 95.7 & 88.2 & 89.0 & 79.0 & 78.9 & 76.9 & 78.9 & 74.8\\
ABD-Net \cite{Chen_2019_ICCV} (ICCV2019) & ResNet-50 & 95.6 & 88.2 & 89.0 & 78.6 & - & - & - & -\\
\hline
DaRe \cite{Wang_2018_CVPR} (CVPR2018) & ResNet-50 & 88.3 & 82.0 & 80.4 & 74.5 & 66.0 & 66.7 & 62.8 & 62.8\\
Consistent-Net \cite{Zhou_2019_ICCV} (ICCV2019) & ResNet-50 & \textbf{96.1} & 84.7 & 86.3 & 73.1 & - & - & - & -\\
BDB+Cut \cite{Dai_2019_ICCV} (ICCV2019) & ResNet-50 & 95.3 & 86.7 & 89.0 & 76.0 & 79.4 & 76.7 & 76.4 & 73.5\\
MHN-6 (PCB) \cite{Chens_2019_ICCV} (ICCV2019) & ResNet-50 & 95.1 & 85.0 & 89.1 & 77.2 & 77.2 & 72.4 & 71.7 & 71.7\\
\hline
HBFP-Net & ResNet-50 & 95.8 & \textbf{89.8} & \textbf{89.5} & \textbf{80.2} & \textbf{81.3} & \textbf{79.4} & \textbf{80.0} & \textbf{77.5}\\
\hline
\end{tabular}
\end{center}
\label{tab1}
\end{table*}
Given an image, we denote $y$ as truth ID label and ${p_i}$ as ID prediction logit of class $i$. In our experiment, we adopt label-smoothed cross-entropy loss \cite{He_2019_CVPR, DBLP:journals/corr/abs-1903-07071} as ID loss function, which is computed as:
\begin{equation}
\label{E8}
{{L_{ID}} = \sum\limits_{i = 1}^C { - {q_i}\log ({p_i})} }
\end{equation}
where $C$ is the number of classes and ${q_i}$ is computed as:
\begin{equation}
\label{E9}
{{q_i}{\rm{ = }}\left\{ {\begin{array}{*{20}{c}}
{1 - \frac{{C - 1}}{C}\varepsilon }\\
{\varepsilon /C}
\end{array}} \right.\begin{array}{*{20}{c}}
{i = y}\\
{{\rm{otherwise}}}
\end{array} }
\end{equation}
where $\varepsilon $ is a small constant to encourage the model to be less confident on the training set, which is set to be 0.3.

\section{Experiment}
In this section, we first introduce three public person Re-ID datasets and evaluation protocols. Then we describe the implementation details and compare the performance of HBFP-Net against existing state-of-the-art methods on the three large-scale datasets. Finally, we conduct ablation study and provide visualization results to illustrate how HBFP-Net achieved its effectiveness.

\subsection{Datasets and Evaluation Metrics}
\textbf{Market1501} \cite{Zheng_2015_ICCV} were captured from 6 different cameras and contains 12,936 training images of 751 identities and 19,732 testing images of 750 identities. The pedestrians are automatically detected by DPM-detector \cite{Pedro_2008_CVPR}. During testing procedure, it contains single-query model and multiple-query model. The single-query model only contains 1 query image of a person and has 3,368 query images. The multiple-query model use the average and max pooling features of multiple images.

\textbf{CUHK03-NP} CUHK03-NP is a new training-testing split protocol for CUHK03 \cite{Li_2014_CVPR}, which contains 14,097 images of 1,467 persons and each identity is observed from two non-overlapping cameras. This dataset has both manually labeled bounding boxes and DPM-detected bounding boxes. We adopt the new training/testing protocol proposed by \cite{Zhong_2017_CVPR}, in which 767 identities are used for training and 700 identities for testing. The labelled dataset includes 7,368 training, 1,400 query and 5,328 gallery images while detected dataset consists of 7,365 training, 1,400 query and 5,332 gallery images.

\textbf{DukeMTMC-ReID} \cite{Ergys_2016_ECCV} is a subset of DukeMTMC specifically collected for person re-identification. It consists of 36,411 images of 1,812 persons from 8 high-resolution cameras, which contains 16,522 training images of 702 identities and a testing set of the remaining identities. The testing set consists of 2,228 query images and 17,661 gallery images. In our experiments, we adopt the training/testing protocol by following \cite{Zheng_2017_ICCV}.

\textbf{Evaluation Metrics}. In our experiments, we employ the
standard cumulative matching characteristics (CMC) accuracy (Rank-1) and the mean average precision(mAP) on all datasets to compare the performance of our proposed method with other SOTA models.

\subsection{Implementation Details}
We choose ResNet-50 as our backbone and initialized from the ImageNet pre-trained model \cite{5206848}. The training images are resized to a resolution of $384 \times 128$ and augmented by random horizontal flip, random erasing, normalization and cutout. The testing images are resized to $384 \times 128$ only with normalization. The spatial size $(H,W)$ and channel number (${D_1}, {D_2}, {D_3}$) of low-, middle- and high-level feature maps are $(24,8)$ and 1024, 1024 and 2048, respectively. The linear mapping $(U, V, U', V', {P_x}, {P_y})$ are all learnable matrix and regularized by Batch Normalization \cite{DBLP:journals/corr/IoffeS15}. The projection dimension $L$ and $L'$ are set to 512 and 2048, respectively. For the triplet loss with hard batch mining \cite{DBLP:journals/corr/HermansBL17}, the margin and the batch size are set to 0.3 and 64 with $P=16$ and $K=4$ to train our model.

We use Adam optimizer \cite{Jimmy_2015_ICLR} with a weight decay of $5 \times {10^{ - 4}}$ to train our model. We set the initial learning rate to 0.0002 and keep it to 200 epochs. Then, the learning rate decays to 0 at exponent rule after 500 epochs. Our model is implemented on Pytorch platform and trained in an end-to-end manner with two NVIDIA GTX-1080Ti GPUs.

\begin{figure*}
\begin{center}
\includegraphics[width=1.0\linewidth]{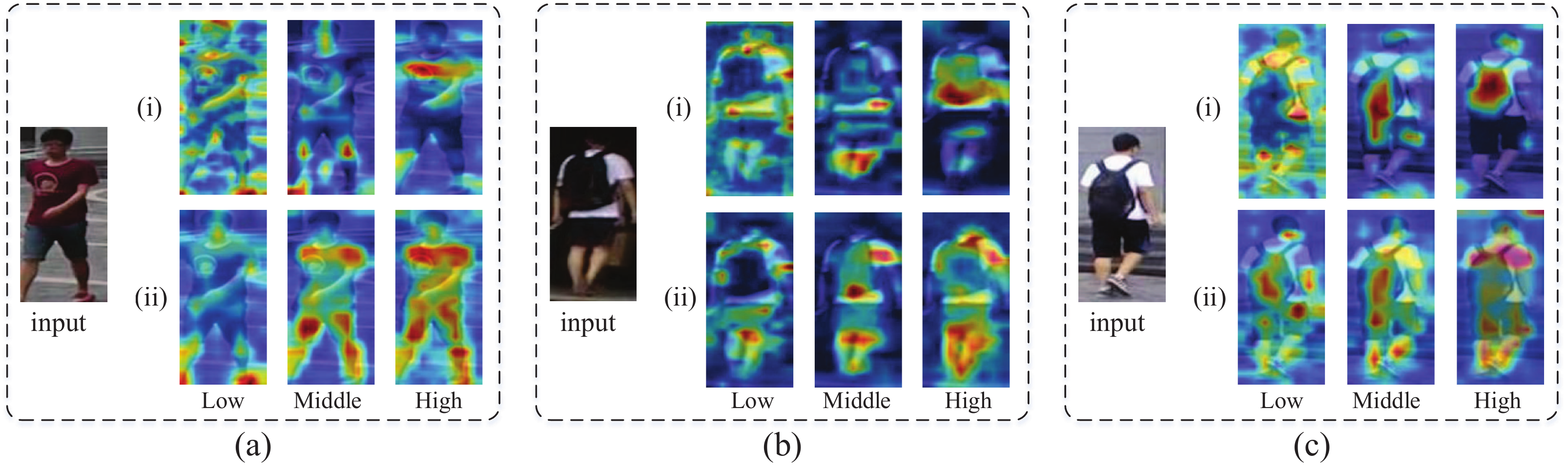}
\end{center}
   \caption{Examples of visualization results of low-, middle- and high-level feature maps with and without BFP. In each example, from left to right are the original image, the visualization of low-, middle- and high-level feature maps. The first row (i) of each example is the visualization of feature maps without BFP and the second row (ii) is the results of corresponding feature maps with BFP.}
\label{fig5}
\end{figure*}

\subsection{Comparison with State-of-the-Art}
We compare the proposed HBFP-Net against the state-of-the-art methods on three widely used large datasets Market-1501, CUHK03 and DukeMTMC-ReID shown in Table~\ref{tab1}, which show that our method achieves superior performance over all comparing methods. We select 6 attention-based methods, 5 partition methods and dropout methods to evaluate the proposed HBFP-Net.
Note that all reported results are obtained without any post-processing techniques such as re-ranking \cite{Zhong_2017_CVPR} or multi-query fusion \cite{Zheng_2017_ICCV}.

Our method achieves 95.8\% rank-1 accuracy and 89.8\% mAP on Market-1501 datasets, which outperforms the typical partition based method PCB \cite{Sun_2018_ECCV} 2\% on rank-1 accuracy and 8.2\% on mAP. Compared with Consistent-Net \cite{Zhou_2019_ICCV}, which achieved the best result on rank-1 accuracy, our method can achieve close performance on rank-1 accuracy (95.8\% VS 96.1\%), but our method exceeds it 5.1\% on mAP(89.8\% VS 84.7\%), which show that the retrieval results of our method are more reliable. SCAL \cite{Chenss_2019_ICCV} integrates channel and spatial attention methods and provides a supervised signal for learned attention mask, which achieved the state-of-the-art results of rank-1/mAP = 95.8\%/89.3\%. Our model achieves comparative performance with it on Market-1501 and DukeMTMC-ReID datasets. However, on CUHK03-labeled and CUHK03-detected datasets, our method shows superior performance (81.3\% VS 74.8\% of rank-1 and 79.4\% VS 72.3\% of mAP in the labeled version; 80.0\% VS 71.1\% of rank-1 and 77.5\% VS 68.6\% of mAP in the detected version).

We conduct experiments on both version of CUHK03-NP benchmark: manually labeled bounding boxes (labeled) and DPM-detected \cite{Pedro_2008_CVPR} bounding boxes (detected). For both version CUHK03-labeled and CUHK03-detected, the proposed HBFP-Net achieves superior performance (81.3\% of rank-1 accuracy and 79.4\% of mAP in the labeled version; 80.0\% of rank-1 accuracy and 77.5\% of mAP in the detected version). In this challenging datasets, our model makes large improvement, which outperforms the state-of-the-art method BDB \cite{Dai_2019_ICCV} 1.9\% of rank-1 accuracy and 2.7\% mAP on CUHK03-label version and 3.6\% of rank-1 accuracy and 4.0\% mAP on CUHK03-detected version. For DukeMTMC-ReID datasets, our proposed method HBFP-Net achieves 89.5\% on rank-1 accuracy and 80.2\% on mAP, which exceeds the BDB \cite{Dai_2019_ICCV} 0.5\% rank-1 accuracy and 4.2\% mAP. What's more, our method surpasses the multi-level attention based method \cite{Zhou_2019_ICCV} 3.2\% on rank-1 accuracy and 7.1\% on mAP. Overall, the results on all three datasets endorse the superiority of the proposed HBFP-Net.

\subsection{Visualizations}
In Fig ~\ref{fig5}, we show three examples of different identities and their CAM \cite{Zhou_2016_CVPR} visualization of the low-, middle- and high-level feature maps with and without BFP. In each example, from left to right are the original image, the visualization of low-, middle- and high-level feature maps. The first row (i) in Fig ~\ref{fig5} represents the original low-, middle- and high-level feature maps. It can be observed that the high-level feature maps without BFP mainly focus on pattern on clothes and backpack respectively, while the information of low-level feature maps is dispersed, which can't capture discriminative semantic regions. The second row (ii) in Fig ~\ref{fig5} shows the augmented feature maps via BFP. It can be seen that the augmented high-level features enlarge attention regions to whole body including head, trousers, shoes and arm, and the information of augmented low-level features is more concentrated, which can capture some important parts like head, arm and leg.

From the visualization of original low-, middle- and high-level feature maps shown in the first row, we can know that the low-level information is too dispersed while the high-level representation is very abstract, so it is hard to directly formulate the relationship between them. We find that the middle-level feature maps is not only abstract but also dispersed, so both high- and low-level feature maps exploit the relationship with middle-level feature to correlate each other, which make the learned correlation maps more reliable.

\subsection{Ablation Study}
To evaluate the effectiveness of each component of the proposed method, we conduct several ablation experiments on Market-1501 dataset in the single query mode. The results are shown in Table~\ref{tab2} and Table~\ref{tab3}, and the settings are the same as BFP implementation detailed in Section 4.2.

\textbf{Baseline}. In Table~\ref{tab2}, ``Baseline'' represents the ResNet-50 only trained with the ranking loss of batch-hard triplet loss \cite{DBLP:journals/corr/HermansBL17}. To further exploit the identity information, we add an another label-smoothed cross-entropy loss\cite{He_2019_CVPR}, described as ``Baseline(+LS)''. It can be observed that the ``Baseline'' and ``Baseline(+LS)'' achieve 89.0\% and 91.3\% in Rank-1 accuracy. Cutout \cite{DBLP:journals/corr/abs-1708-04552} is a simple and effect data augmentation technique, which randomly masks out square regions of input images during training procedure. When adding cutout technique in input images, the result can achieve 92.8\% rank-1 accuracy and 83.0\% of mAP, represented as ``Baseline(+LS+Cutout)'' in Table~\ref{tab2}. Note that our following experiments are based on ``Baseline(+LS+Cutout)'' and regard it as our baseline.

\begin{table}
\caption{Ablation study of our method on Market-1501 dataset. GeP represents generalized pooling.}
\begin{center}
\begin{tabular}{c|ccc}
\hline
Method & mAP & Rank-1 & Rank-5 \\
\hline
Baseline & 75.2 & 89.0 & 93.4  \\
Baseline(+LS) & 81.2 & 91.3 & 96.5  \\
Baseline(+LS+Cutout) & 83.0 & 92.8 & 97.2  \\
\hline
Middle+High(Res) & 85.2 & 93.1 & 98.2  \\
Low+Middle+High(Res) & 77.2 & 91.8 & 96.7  \\
\hline
BFP(Low\&Middle)+High & 86.5 & 94.1 & 98.2  \\
BFP(Middle\&High) & 87.2 & 94.2 & 98.3 \\
Low+BFP(Middle\&High) & 84.2 & 93.1 & 97.0 \\
BFP(Low\&Middle\&High) & 88.0 & 94.8 & 98.5 \\
HBFP & 88.7 & 95.0 & 98.5 \\
HBFP+GeP & 89.8 & 95.8 & 98.9 \\
\hline
\end{tabular}
\end{center}
\label{tab2}
\end{table}

\begin{table}
\caption{Impact of the parameter $\lambda $ of generalized pooling.}
\begin{center}
\begin{tabular}{c|ccc}
\hline
GeP & mAP & Rank-1 & Rank-5 \\
\hline
$\lambda {\rm{ = }}0$ & 88.7 & 95.0 & 98.5 \\
+ $(\lambda {\rm{ = }}0.3)$ & 89.0 & 95.2 & 98.6 \\
+ $(\lambda {\rm{ = }}0.5)$ & 89.5 & 95.5 & 98.8 \\
+ $(\lambda {\rm{ = }}0.7)$ & 89.6 & 95.6 & 98.7 \\
+ $(\lambda {\rm{ = }}1.0)$ & 89.8 & 95.8 & 98.9 \\
\hline
\end{tabular}
\end{center}
\label{tab3}
\end{table}

\textbf{The effect of BFP}. To verify that the improvement comes from BFP, we conduct two comparison experiments which learn low-, middle- and high-level features with or without BFP module. First, we use an intuitional model to combine middle- and high-level information, which trains middle- and high-level features respectively without BFP moduel, represented as ``Middle + High(Res)'' in Table~\ref{tab2}. It can be observed that its performance achieve improvements of 0.3\% on rank-1 and 2.2\% on mAP. However, when combination of low-level feature, represented as ``Low + Middle + High(Res)'', rank-1 accuracy and mAP decrease to 91.8\% and 77.2\%, which demonstrate that learning low-level feature individually plays a negative influence during retrieval process. After employing BFP module on middle- and high-level features, shown as ``BFP(Middle \& High)'', its performance achieves 94.2\% of rank-1 accuracy and 87.2\% of mAP. And when combination of low-level feature without BFP, its result decrease to 93.1\% of rank-1 accuracy and 84.2\% of mAP, represented as ``Low + BFP(Middle \& High)''. ``BFP(Low \& Middle) + High'' in Table~\ref{tab2} represents the BFP module employed on low- and middle-level features and high-level feature learned individually, which achieves 94.1\% of rank-1 accuracy and 86.5\% of mAP. After employing BFP module on low-, middle- and high-level features, it can achieve 94.8\% of rank-1 accuracy and 88.0\% of mAP. What's more, when combination of generalized pooling, represented as ``HBFP + GeP'' in Table~\ref{tab2}, it achieves the best result of 95.8\% on rank-1 accuracy and 89.8\% on mAP.

\textbf{The effect of generalized pooling}. Since the augmented features with BFP enlarge the attention regions to whole body, we employ generalized pooling to capture gradual cues by different setting of $\lambda $ in Eq.~(\ref{E6}), and the influence of different $\lambda $ is shown in Table~\ref{tab3}. It can be known that $\lambda {\rm{ = }}0$ represents average pooling, which achieves 95.0\% of rank-1 accuracy and 88.7\% of mAP. To capture gradual cues, we set the interval of 0.2. When combination of $\lambda {\rm{ = }}0.3$ and $\lambda {\rm{ = }}0.5$, the results improve to 95.5\% on rank-1 accuracy and 89.5\% on mAP. To further capture salient cues, we combine $\lambda {\rm{ = }}0.7$ and $\lambda {\rm{ = }}1.0$, which achieves 95.8\% of rank-1 accuracy and 89.8\% of mAP. Actually, $\lambda {\rm{ = }}1.0$ represents the max pooling. When $\lambda $ is set to 0, it preserves all values even background and $\lambda {\rm{ = }}1.0$ will select the largest response values of all spatial locations. We combine $\lambda {\rm{ = }}0.3$, $\lambda {\rm{ = }}0.5$ and $\lambda {\rm{ = }}0.7$ to capture the gradual cues. So in our model, we choose five gradual cues of $\lambda {\rm{ = }}0$, $\lambda {\rm{ = }}0.3$, $\lambda {\rm{ = }}0.5$, $\lambda {\rm{ = }}0.7$ and $\lambda {\rm{ = }}1.0$ to extract features and sum them as the final representation of an image.

\section{Conclusion}
In this paper, we propose a novel Hierarchical Bi-directional Feature Perception Network (HBFP-Net) to enrich the attention regions of high-level feature, and to learn abstract and specific information in low-level feature that in turn benefits the deep network. The correlation maps of cross-level features are modeled via low-rank bilinear pooling and used for correlate each other through our proposed Bi-directional Feature Perception module. To capture gradual cues of feature maps, a novel trainable generalized pooling is leveraged. Extensive experiments on three challenging datasets show that HBFP-Net achieves superior performance and outperforms the state-of-the arts.



\bibliographystyle{ACM-Reference-Format}
\bibliography{sample-base}

\appendix

\end{document}